\documentclass[article,shortnames,nojss]{jss}

\usepackage{thumbpdf} 
\usepackage{amsmath,amssymb}
\usepackage{tikz}
\usepackage{float}
\newfloat{algorithm}{tbp}{loa}
\providecommand{\algorithmname}{Algorithm}
\floatname{algorithm}{\protect\algorithmname}

\title{\pkg{copulaedas}: An \proglang{R} Package for Estimation of Distribution Algorithms Based on Copulas}
\Plaintitle{copulaedas: An R Package for Estimation of Distribution Algorithms Based on Copulas}
\Shorttitle{\pkg{copulaedas}: An \proglang{R} Package for EDAs Based on Copulas}

\Abstract{
The use of copula-based models in EDAs (estimation of distribution algorithms) 
is currently an active area of research. In this context, the \pkg{copulaedas} 
package for \proglang{R} provides a platform where EDAs based on copulas can 
be implemented and studied. The package offers complete implementations of 
various EDAs based on copulas and vines, a group of well-known optimization 
problems, and utility functions to study the performance of the algorithms. 
Newly developed EDAs can be easily integrated into the package by extending 
an \proglang{S}4 class with generic functions for their main components. 
This paper presents \pkg{copulaedas} by providing an overview of EDAs based
on copulas, a description of the implementation of the package, and an 
illustration of its use through examples. The examples include running the 
EDAs defined in the package, implementing new algorithms, and performing 
an empirical study to compare the behavior of different algorithms on 
benchmark functions and a real-world problem.
}

\Keywords{black-box optimization, estimation of distribution algorithm, copula, vine, \proglang{R}}
\Plainkeywords{black-box optimization, estimation of distribution algorithm, copula, vine, R}

\author{
Yasser Gonzalez-Fernandez \\ 
Institute of Cybernetics, \\
Mathematics and Physics 
\And
Marta Soto \\
Institute of Cybernetics, \\
Mathematics and Physics
}
\Plainauthor{Yasser Gonzalez-Fernandez, Marta Soto}

\Address{
Yasser Gonzalez-Fernandez \\
Department of Interdisciplinary Mathematics \\
Institute of Cybernetics, Mathematics and Physics \\
Calle 15 No. 551 e/ C y D, Vedado, La Habana, Cuba \\
\emph{Current affiliation:} \\
School of Information Technology \\
York University \\
4700 Keele St, Toronto, Ontario, Canada \\
E-mail: \email{ygf@yorku.ca} \\

Marta Soto \\
Department of Interdisciplinary Mathematics \\
Institute of Cybernetics, Mathematics and Physics \\
Calle 15 No. 551 e/ C y D, Vedado, La Habana, Cuba \\
E-mail: \email{mrosa@icimaf.cu}
}

\begin{document}

\section{Introduction}

The field of numerical optimization (see e.g., \citealp{Nocedal1999NumericalOptimization})
is a research area with a considerable number of applications in engineering, 
science, and business. Many mathematical problems involve finding the most 
favorable configuration of a set of parameters that achieve an objective quantified 
by a function. Numerical optimization entails the case where these parameters 
can take continuous values, in contrast with combinatorial optimization, 
which involves discrete variables. The mathematical formulation of a numerical 
optimization problem is given by $\underset{\boldsymbol{x}}{\min\,}f(\boldsymbol{x})$,
where~$\boldsymbol{x}\in\mathbb{R}^{n}$ is a real vector with $n\geq1$~components
and $f:\mathbb{R}^{n}\rightarrow\mathbb{R}$ is the objective function
(also known as the fitness, loss or cost function).

In particular, we consider within numerical optimization a black-box
(or direct search) scenario where the function values of evaluated search
points are the only available information on~$f$. The algorithms do 
not assume any knowledge of the function~$f$ regarding continuity,
the existence of derivatives, etc. A black-box optimization procedure 
explores the search space by generating solutions, evaluating them, and 
processing the results of this evaluation in order to generate new promising 
solutions. In this context, the performance measure of the algorithms is 
generally the number of function evaluations needed to reach a certain 
value of~$f$.

Algorithms that have been proposed to deal with this kind of optimization
problems can be classified in two groups according to the approach
followed for the generation of new solutions. On the one hand, deterministic
direct search algorithms, such as the Hooke-Jeeves \citep{Hooke1961HookeJeeves}
and Nelder-Mead \citep{Nelder1965Simplex} methods, perform transformations
to one or more candidate solutions at each iteration. Given their
deterministic approach, these algorithms may have limited global search
capabilities and can get stuck in local optima, depending on an appropriate
selection of the initial solutions. On the other hand, randomized
optimization algorithms offer an alternative to ensure a proper global
exploration of the search space. Examples of these algorithms are
simulated annealing \citep{Kirkpatrick1983SA}, evolution strategies
(see e.g., \citealp{Beyer2002ESIntro}), particle swarm optimization
\citep{Kennedy1995PSO}, and differential evolution \citep{Storn1997DE}.

In this paper, we focus on EDAs \citep[estimation of distribution
algorithms;][]{Muhlenbein1996BinaryParameters,Baluja1994PBIL,Larranaga2002EDANewToolEC,Pelikan2002SurveyOptimization},
which are stochastic black-box optimization algorithms characterized
by the explicit use of probabilistic models to explore the search
space. These algorithms combine ideas from genetic and evolutionary
computation, machine learning, and statistics into an optimization
procedure. The search space is explored by iteratively estimating and
sampling from a probability distribution built from promising
solutions, a characteristic that differentiates EDAs among other
randomized optimization algorithms. One key advantage of EDAs is that
the search distribution may encode probabilistic dependences between
the problem variables that represent structural properties of the
objective function, performing a more effective optimization by using
this information.

Due to its tractable properties, the normal distribution has been
commonly used to model the search distributions in EDAs for real-valued
optimization problems \citep{Bosman2006RealValuedEDAs,Kern2003ReviewContinuousEDAs}.
However, once a multivariate normal distribution is assumed, all the
margins are modeled with the normal density and only linear correlation
between the variables can be considered. These characteristics could
lead to the construction of incorrect models of the search space.
For instance, the multivariate normal distribution cannot represent
properly the fitness landscape of multimodal objective functions.
Also, the use of normal margins imposes limitations on the performance
when the sample of the initial solutions is generated asymmetrically
with respect to the optimum of the function (see \citealp{Soto2014CopulaVineEDA}
for an illustrative example of this situation).

Copula functions (see e.g.,
\citealt{Joe1997MultivariateModels,Nelsen2006IntroductionCopulas})
offer a valuable alternative to tackle these problems. By means of
Sklar's Theorem \citep{Sklar1959CopulasFrench}, any multivariate
distribution can be decomposed into the (possibly different)
univariate marginal distributions and a multivariate copula that
determines the dependence structure between the variables. EDAs based
on copulas inherit these properties, and consequently, can build more
flexible search distributions that may overcome the limitations of a
multivariate normal probabilistic model. The advantages of using
copula-based search distributions in EDAs extend further with the
possibility of factorizing the multivariate copula with the copula
decomposition in terms of lower-dimensional copulas. Multivariate
dependence models based on copula factorizations, such as nested
Archimedean copulas \citep{Joe1997MultivariateModels} and vines
\citep{Joe1996hFunctions,Bedford2001DensityDecomposition,Aas2009PairCopulaConstructions},
provide great advantages in high dimensions. Particularly in the case
of vines, a more appropriate representation of multivariate
distributions having pairs of variables with different types of
dependence is possible.

Although various EDAs based on copulas have been proposed in the
literature, as far as we know there are no publicly available
implementations of these algorithms (see
\citealp{Santana2011EDAImplementations} for a comprehensive review of
EDA software). Aiming to fill this gap, the \pkg{copulaedas} package
\citep{copulaedasRPackage} for the \proglang{R} language and
environment for statistical computing \citep{RCoreTeam2013R} has been
published on the Comprehensive \proglang{R} Archive Network at
\url{http://CRAN.R-project.org/package=copulaedas}. This package
provides a modular platform where EDAs based on copulas can be
implemented and studied. It contains various EDAs based on copulas, a
group of well-known benchmark problems, and utility functions to study
EDAs. One of the most remarkable features of the framework offered by
\pkg{copulaedas} is that the components of the EDAs are decoupled 
into separated generic functions, which promotes code factorization 
and facilitates the implementation of new EDAs that can be easily 
integrated into the framework.

The remainder of this paper provides a presentation of the \pkg{copulaedas}
package organized as follows. Section~\ref{sec:copulaedas} continues
with the necessary background on EDAs based on copulas. Next, the
details of the implementation of \pkg{copulaedas} are described in
Section~\ref{sec:implementation}, followed by an illustration of
the use of the package through examples in Section~\ref{sec:examples}.
Finally, concluding remarks are given in Section~\ref{sec:conclusions}.

\section[EDAs based on copulas]{Estimation of distribution algorithms based on copulas\label{sec:copulaedas}}

This section begins by describing the general procedure of an EDA,
according to the implementation in \pkg{copulaedas}. Then, we present
an overview of the EDAs based on copulas proposed in the literature
with emphasis on the algorithms implemented in the package.

\subsection{General procedure of an EDA}

The procedure of an EDA is built around the concept of performing
the repeated refinement of a probabilistic model that represents the
best solutions of the optimization problem. A typical EDA starts with
the generation of a population of initial solutions sampled from the
uniform distribution over the admissible search space of the problem.
This population is ranked according to the value of the objective
function and a subpopulation with the best solutions is selected.
The algorithm then constructs a probabilistic model to represent the
solutions in the selected population and new offspring are generated
by sampling the distribution encoded in the model. This process is
repeated until some termination criterion is satisfied (e.g., when a
sufficiently good solution is found) and each iteration of this procedure 
is called a generation of the EDA. Therefore, the feedback for the 
refinement of the probabilistic model comes from the best solutions 
sampled from an earlier probabilistic model.

\begin{figure}[p!]
\centering
\input{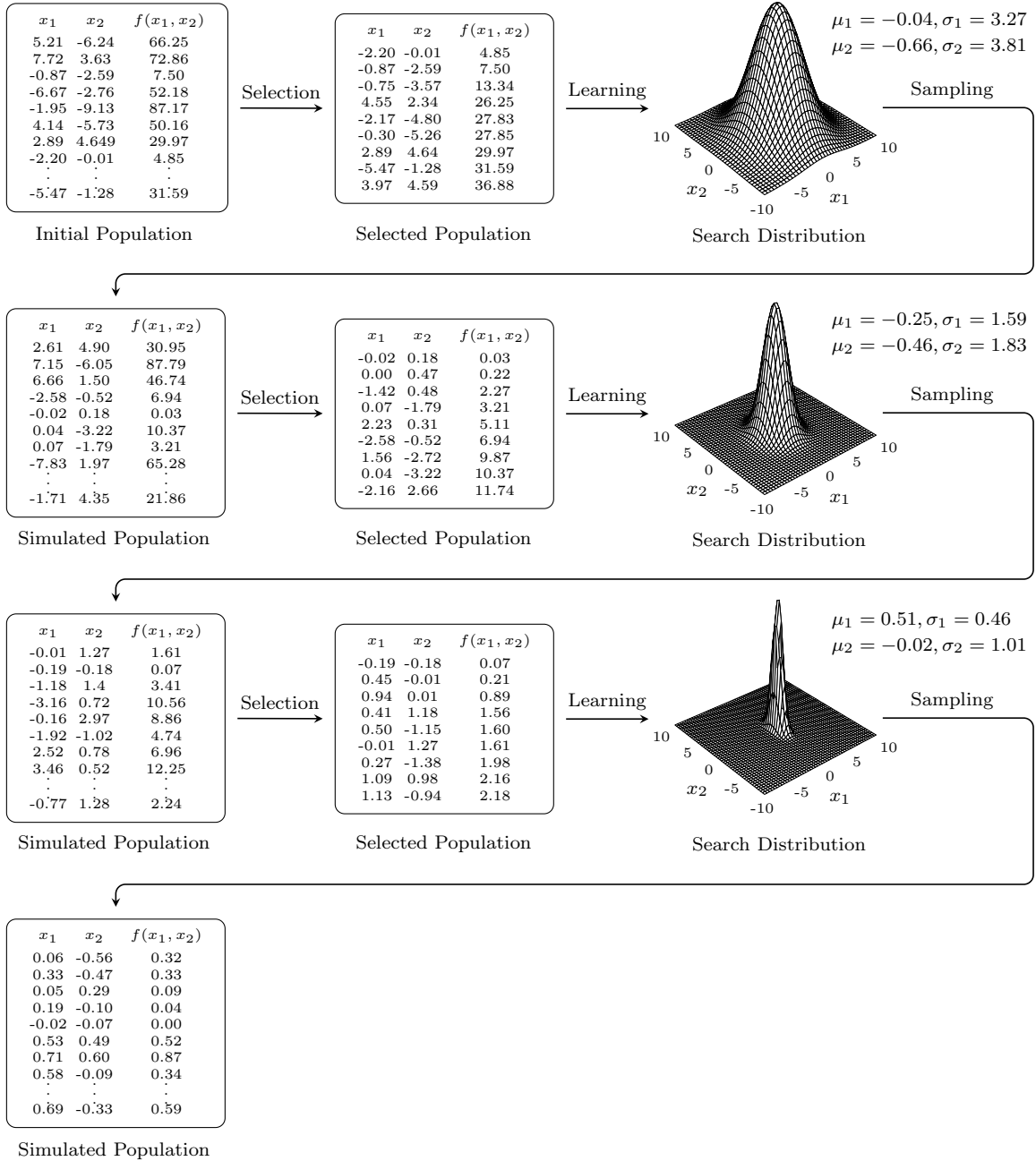}
\par
\caption{Steps performed to minimize the function $f(x_{1},x_{2})=x_{1}^{2}+x_{2}^{2}$
using a continuous EDA that assumes independence between the variables.
The search distribution models each variable with the normal distribution
and, since mutual independence is assumed, the joint PDF is factorized
as $\phi_{1,2}(x_{1},x_{2})=\phi_{1}(x_{1},\mu_{1},\sigma_{1}^{2})\phi_{2}(x_{2},\mu_{2},\sigma_{2}^{2})$.
The simulation of the probabilistic model estimated at the third generation
produces a solution $f(-0.02,-0.07)=0.00$ that approximates the global
optimum of the function with the required precision.\label{fig:eda-example}}
\end{figure}

Let us illustrate the basic EDA procedure with a concrete example.
Figure~\ref{fig:eda-example} shows the steps performed to minimize
the two-dimensional objective function $f(x_{1},x_{2})=x_{1}^{2}+x_{2}^{2}$
using a simple continuous EDA that assumes independence between the
problem variables. Specifically, we aim to find the global optimum
of the function $f(0,0)=0$ with a precision of two decimal places.

The algorithm starts by generating an initial population of 30
candidate solutions from a continuous uniform distribution in
$[-10,10]^{2}$.  Out of this initial sampling, the best solution found
so far is $f(-2.20,-0.01)=4.85$.  Next, the initial population is
ranked according to their evaluation in $f(x_{1},x_{2})$, and the best
30\% of the solutions is selected to estimate the probabilistic
model. This EDA factorizes the joint probability density function
(PDF) of the best solutions as
$\phi_{1,2}(x_{1},x_{2})=\phi_{1}(x_{1},\mu_{1},\sigma_{1}^{2})\phi_{2}(x_{2},\mu_{2},\sigma_{2}^{2})$,
which describes mutual independence, and where $\phi_{1}$ denotes the
univariate normal PDF of $x_{1}$ with mean $\mu_{1}$ and variance
$\sigma_{1}^{2}$, and $\phi_{1}$ denotes the univariate normal PDF of
$x_{2}$ with mean $\mu_{1}$ and variance $\sigma_{2}^{2}$. In the
first generation, the parameters of the probabilistic model are
$\mu_{1}=-0.04$, $\sigma_{1}=3.27$, $\mu_{2}=-0.66$ and
$\sigma_{2}=3.81$. The second generation starts with the simulation of
a new population from the estimated probabilistic model. Afterwards,
the same selection procedure is repeated and the resulting selected
population is used to learn a new probabilistic model. These steps are
then repeated for a third generation.

Notice how in the first three generations the refinement of the probabilistic
model that represents the best solutions is evidenced in the reduction
of the variance of the marginal distributions towards a mean value
around zero. Also, the convergence of the algorithm is reflected in
the reduction of the value of the objective function from one generation
to another. Ultimately, the simulation of the probabilistic model
estimated at the third generation produces $f(-0.02,-0.07)=0.00$,
which satisfies our requirements and the algorithm terminates.

In practice, EDAs include other steps in addition to the ones illustrated
in the previous basic example. The general procedure of an EDA implemented
in \pkg{copulaedas} is outlined in Algorithm~\ref{alg:eda-procedure}.
In the following, we provide a description of the purpose of the main
steps of the algorithm, which are highlighted in italics in the pseudocode:

\begin{algorithm}
\centering
\begin{tabular}{l}
\hline
$i \leftarrow 1$ \\
\textbf{repeat} \\
\hspace{1em} \textbf{if} $i = 1$ \textbf{then} \\
\hspace{2em} Generate an initial population $P_1$ 
             using a \emph{seeding method}. \\
\hspace{2em} Evaluate the solutions in the population $P_1$. \\
\hspace{2em} If required, apply a \emph{local optimization method} 
             to the population $P_1$. \\
\hspace{1em} \textbf{else} \\
\hspace{2em} Select a population $P_i^{Selected}$ from $P_{i-1}$
             according to a \emph{selection method}. \\
\hspace{2em} Learn a probabilistic model $M_i$ from $P_i^{Selected}$
             using a \emph{learning method}.  \\
\hspace{2em} Sample a new population $P_i^{Sampled}$ from $M_i$
             using a \emph{sampling method}.  \\
\hspace{2em} Evaluate the solutions in the population $P_i^{Sampled}$. \\
\hspace{2em} If required, apply a \emph{local optimization method} 
             to the population $P_i^{Sampled}$. \\
\hspace{2em} Create the population $P_i$ from $P_{i-1}$ and 
             $P_i^{Sampled}$ using a \emph{replacement method}. \\
\hspace{1em} \textbf{end if} \\
\hspace{1em} If required, report progress information using a 
             \emph{reporting method}. \\
\hspace{1em} $i \leftarrow i + 1$ \\
\textbf{until} A criterion of the \emph{termination method} is met. \\
\hline
\end{tabular}
\caption{Pseudocode of an EDA.\label{alg:eda-procedure}}
\end{algorithm}

\begin{itemize}
\item The first step is the generation of an initial population of solutions
following a \emph{seeding method}, which is usually random, but it
can use a particular heuristic when a priori information about the
characteristics of the problem is available. 

\item The results of global optimization algorithms such as EDAs can often
be improved if combined with \emph{local optimization methods} that
look for better solutions in the neighborhood of each candidate solution.
 Local optimization methods can also be used to implement repairing
methods for constrained problems where the simulated solutions may
be unfeasible and a strategy to repair these solutions is available.

\item A \emph{selection method} is used to determine the most promising
solutions of the population. An example selection method is truncation
selection, which creates the selected population with a percentage
of the best solutions of the current population.

\item The estimation and simulation of the search distribution are the essential
steps of an EDA. These steps are implemented by \emph{learning and
sampling methods}, which are tightly related. Learning methods estimate
the structure and parameters of the probabilistic model used by the algorithm
from the selected population, while sampling methods are used to generate
new solutions from the estimated probabilistic model.

\item A \emph{replacement method} is used to incorporate a new group of
solutions into the current population. For example, a replacement
strategy is to substitute the current population with the newly sampled
population. Other replacement strategies retain the best solutions
found so far or try to maintain the diversity of solutions.

\item \emph{Reporting methods} provide progress information during the execution
of the EDA. Relevant progress information can be the number of evaluations
of the objective function and the best solution found so far.

\item A \emph{termination method} determines when the algorithm stops according
to certain criteria; for example, when a fixed number of function
evaluations are realized or a certain value of the objective function
is reached.
\end{itemize}

Although it was possible to locate with the required precision the
optimum of the simple function presented in this section, it is not
always possible to perform a successful search by considering only the
marginal information. As we show later in this paper, the assumption
of independence between the variables constitutes a strong limitation
that may compromise the convergence of an EDA. The use of information
about the relationships between the variables allows searching efficiently 
for the best solutions and it constitutes one of the main advantages of EDAs.
Among the algorithms that consider dependences between the variables, 
we are particularly interested in EDAs whose learning and sampling 
steps involve probabilistic models based on copulas. The next section 
provides an overview of such algorithms.

\subsection{Overview of EDAs based on copulas}

To the best of our knowledge, the technical report \citep{Soto2007EDACopulaGaussiana}
and the theses \citep{Arderi2007BachelorThesis,Barba-Moreno2007MastersThesis}
constitute the first attempts to incorporate copulas into EDAs. Since then,
a considerable number of EDAs based on copula theory have been proposed
in the literature and, as evidence of its increasing popularity, the
use of copulas in EDAs has been identified as an emerging approach
for the solution of real-valued optimization problems \citep{Hauschild2011EDASurveyPaper}.

The learning step of copula-based EDAs consists of two tasks: the
estimation of the marginal distributions and the estimation of the
probabilistic dependence structure. In general, these tasks have been
performed by following one of the two-step estimation procedures known
in the copula literature as the IFM \citep[inference functions for
margins;][]{Joe1996IFMEstimation,Joe2005EfficiencyIFM} and the
semiparametric estimation method
\citep{Genest1995SemiparametricEstimation}. Firstly, the marginal
distributions are estimated and the selected population is transformed
into uniformly distributed variables in $(0,1)$ by means of the
evaluation of each marginal cumulative distribution
function. Secondly, the transformed population is used to estimate a
copula-based model of the dependence structure among the
variables. Usually, a particular parametric distribution (e.g., normal
or beta) is assumed for each margin and its parameters are estimated
by maximum likelihood (see e.g.,
\citealt{Soto2007EDACopulaGaussiana,Salinas-Gutierrez2009CopulasEDAs}).
In other cases, empirical marginal distributions or kernel density
estimation have been used (see e.g.,
\citealt{Soto2007EDACopulaGaussiana,Gao2009EDALaplaceTransform,Cuesta-Infante2010ArchimedeanCopulasEDA}).
The simulation step typically starts with the generation of a
population of uniformly distributed variables in $(0,1)$ with the
dependence structure described by the copula-based model that was
estimated in the learning step.  Finally, this uniform population is
transformed to the domain of the variables through the evaluation of
the inverse of each marginal cumulative distribution function.

According to the copula model being used, EDAs based on copulas can
be classified as EDAs based on either multivariate or factorized copulas.
In the rest of this section we give an overall description of representative
algorithms belonging to each group that have been proposed in the
literature.

\subsubsection{EDAs based on multivariate copulas}

The research on EDAs based on multivariate copulas has focused on the
use of multivariate elliptical copulas
\citep{Abdous2005EllipticalDistributions,Fang2002MetaElliptical} and
Archimedean copulas
\citep{Joe1997MultivariateModels,McNeil2009ArchimedeanCopulas}.  The
algorithms described in
\cite{Soto2007EDACopulaGaussiana,Arderi2007BachelorThesis} and
\cite{Barba-Moreno2007MastersThesis} are both based on the
multivariate normal copula and theoretically similar, but they present
differences in the estimation of the marginal distributions and the
use of techniques such as variance scaling.  \citet{Wang2009EDACopula}
present an EDA based on the bivariate normal copula and, since only
normal marginal distributions are used, the proposed algorithm is
equivalent to EMNA \citep[estimation of multivariate normal
algorithm;][]{Larranaga2001NormalGaussianNetworks}. On the other hand,
the algorithms presented in \cite{Wang2009EDAArchimedianCopulas} and
\cite{Gao2009EDALaplaceTransform} use exchangeable Archimedean
copulas. \citet{Wang2009EDAArchimedianCopulas} propose two algorithms
that use Clayton and Ali-Mikhail-Haq copulas with fixed parameters,
while \citet{Gao2009EDALaplaceTransform} does not state which
particular members of the family of Archimedean copulas are used.

Two EDAs based on multivariate copulas are implemented in
\pkg{copulaedas}, one is based on the product or independence copula
and the other on the normal copula. The first algorithm is UMDA
(univariate marginal distribution algorithm) for continuous variables
\citep{Larranaga1999GaussianMIMIC,Larranaga2000GaussianNetworks},
which can be integrated into the framework of copula-based EDAs
although originally it was not defined in terms of copulas. A
consequence of Sklar's Theorem is that random variables are
independent if and only if the underlying copula is the product
copula. Thus, UMDA can be described as an EDA that models the
dependence structure between the variables using a multivariate
product copula. 

The second EDA based on a multivariate copula implemented in 
\pkg{copulaedas} is GCEDA \citep[Gaussian copula estimation of distribution
algorithm;][]{Soto2007EDACopulaGaussiana,Arderi2007BachelorThesis}.
This algorithm is based on the multivariate normal copula, which
allows the construction of multivariate distributions with normal
dependence structure and non-normal margins. The dependence structure
of the multivariate normal copula is determined by a positive-definite
correlation matrix. If the marginal distributions are not normal, the
correlation matrix is estimated through the inversion of the
non-parametric estimator of Kendall's tau for each pair of variables
(see e.g., \citealt{Genest2007CopulaAfraidAsk,Hult2002TauRhoNormalCopula}). 
If the resulting matrix is not positive-definite, the transformation
proposed by \citet{Rousseeuw1993TransformCorrMatrix} is applied.
GCEDA is equivalent to EMNA when all the marginal distributions are
normal.

\subsubsection{EDAs based on copula factorizations}

The use of multivariate copulas to model the dependence structure
between the variables offers a considerable number of advantages over
the use of the multivariate normal distribution; nevertheless, it
presents limitations. The number of tractable copulas available when
more than two variables are involved is limited, most available
copulas are just investigated in the bivariate case. In addition, the
multivariate elliptical copulas might not be appropriate when all
pairs of variables do not have the same dependence structure. Another
limitation is that some multivariate extensions, such as exchangeable
Archimedean copulas or the multivariate $t$~copula, have only one
parameter to describe certain aspects of the overall dependence. This
characteristic can be a serious limitation when the type and strength
of the dependence is not the same for all pairs of variables. One
alternative to these limitations is to use copula factorizations that
build high-dimensional probabilistic models by using lower-dimensional
copulas as building blocks. Several EDAs based on copula
factorizations, such as nested Archimedean copulas
\citep{Joe1997MultivariateModels} and vines
\citep{Joe1996hFunctions,Bedford2001DensityDecomposition,Aas2009PairCopulaConstructions},
have been proposed in the literature.

The EDA introduced in \cite{Salinas-Gutierrez2009CopulasEDAs} is an
extension of MIMIC (mutual information maximization for input
clustering) for continuous domains
\citep{Larranaga1999GaussianMIMIC,Larranaga2000GaussianNetworks} that
uses bivariate copulas in a chain structure instead of bivariate
normal distributions. Two instances of this algorithm were presented,
one uses normal copulas and the other Frank copulas. In
Section~\ref{sec:example-new-eda}, we illustrate the implementation of
this algorithm using \pkg{copulaedas}.

The exchangeable Archimedean copulas employed in
\cite{Wang2009EDAArchimedianCopulas} and
\cite{Gao2009EDALaplaceTransform} represent highly specialized
dependence structures
\citep{Berg2007ModelsMultivariateDependence,McNeil2008SamplingNAC}.
Within the domain of Archimedean copulas, nested Archimedean copulas
provide a more flexible alternative to build multivariate copula
distributions.  In particular, hierarchically nested Archimedean
copulas present one of the most flexible solutions among the different
nesting structures that have been studied (see e.g.,
\citealt{Berg2007ModelsMultivariateDependence} for a review).
Building from these models, \citet{Ye2010EDANAC} propose an EDA that
uses a representation of hierarchically nested Archimedean copulas
based on L\'evy subordinators \citep{Hering2010HACLevySubordinators}.

\citet{Cuesta-Infante2010ArchimedeanCopulasEDA} investigate the use
of bivariate empirical copulas and a multivariate extension of Archimedean
copulas. The EDA based on bivariate empirical copulas is completely
nonparametric: it employs empirical marginal distributions and a construction
based on bivariate empirical copulas to represent the dependence between
the variables. The marginal distributions and the bivariate empirical
copulas are defined through the linear interpolation of the sample
in the selected population. The EDA based on Archimedean copulas uses
a construction similar to a fully nested Archimedean copula and uses
copulas from one of the families Frank, Clayton or HRT (i.e., heavy
right tail copula or Clayton survival copula). The parameters of the
copulas are fixed to a constant value, i.e., not estimated from the
selected population. The marginal distributions are modeled as in
the EDA based on bivariate empirical copulas.

The class of VEDAs (vine EDAs) is introduced in \cite{Soto2010VEDA}
and \cite{Gonzalez-Fernandez2011BachelorThesis}.  Algorithms of this
class model the search distributions using regular vines, which are
graphical models that represent a multivariate distribution by
decomposing the corresponding multivariate density into conditional
bivariate copulas, unconditional bivariate copulas and univariate
densities. In particular, VEDAs are based on the simplified
pair-copula construction \citep{HobaekHaff2010SimplifiedPCC}, which
assumes that the bivariate copulas depend on the conditioning
variables only through their arguments. Since all bivariate copulas do
not have to belong to the same family, regular vines model a rich
variety of dependences by combining bivariate copulas from different
families.

A regular vine on $n$~variables is a set of nested trees
$T_{1},\ldots,T_{n-1}$, where the edges of tree~$T_{j}$ are the nodes
of the tree~$T_{j+1}$ with $j=1,\ldots,n-2$. The edges of the trees
represent the bivariate copulas in the decomposition and the nodes
their arguments. Moreover, the proximity condition requires that two
nodes in tree~$T_{j+1}$ are joined by an edge only if the
corresponding edges in~$T_{j}$ share a common node. C-vines (canonical
vines) and D-vines (drawable vines) are two particular types of
regular vines, each of which determines a specific decomposition of
the multivariate density. In a C-vine, each tree~$T_{j}$ has a unique
root node that is connected to $n-j$~edges.  In a D-vine, no node is
connected to more than two edges. Two EDAs based on regular vines are
presented in \cite{Soto2010VEDA} and
\cite{Gonzalez-Fernandez2011BachelorThesis}: CVEDA (C-vine EDA) and
DVEDA (D-vine EDA) based on C-vines and D-vines, respectively. Since
both algorithms are implemented in \pkg{copulaedas}, we describe them
in more detail in the rest of this section.

The general idea of the simulation and inference methods for C-vines
and D-vines was developed by \citet{Aas2009PairCopulaConstructions}.
The simulation algorithm is based on the conditional distribution
method (see e.g., \citealt{Devroye1986RandomVariateGeneration}), while
the inference method should consider two main aspects: the selection
of the structure of the vines and the choice of the bivariate copulas.
In the rest of this section we describe how these aspects are performed
in the particular implementation of CVEDA and DVEDA.

The selection of the structure of C-vines and D-vines is restricted 
to the selection of the bivariate dependences explicitly modeled 
in the first tree. This is accomplished by using greedy heuristics, 
which use the empirical Kendall's tau assigned to the edges of 
the tree. In a C-vine, the node that maximizes the sum of
the weights of its edges to the other nodes is chosen as the root of 
the first tree and a canonical root node is assumed for the rest of
the trees. In a D-vine, the construction of the first tree consists 
of finding the maximum weighted sequence of the variables, which 
can be transformed into a TSP (traveling salesman problem)
instance \citep{Brechmann2010DiplomaThesis}.  For efficiency reasons,
in \pkg{copulaedas} we find an approximate solution of the TSP by
using the cheapest insertion heuristic
\citep{Rosenkrantz1977TSPHeuristics}.

The selection of each bivariate copula in both CVEDA and DVEDA starts
with an independence test
\citep{Genest2004TestIndepEmpiricalCopula,Genest2007CvMIndepTests}.
The product copula is selected when there is not enough evidence
against the null hypothesis of independence at a given significance
level.  Otherwise, the parameters of a group of candidate copulas are
estimated and the copula that minimizes a Cram\'er-von Mises statistic
of the empirical copula is selected
\citep{Genest2008ValidityParametricGoF}.

The cost of the construction of C-vines and D-vines increases with the
number of variables. To reduce this cost, we apply the truncation
strategy presented in \cite{Brechmann2010DiplomaThesis}, for which the
theoretical justification can be found in
\cite{Joe2010VineTailDependence}.  When a vine is truncated at a
certain tree during the tree-wise estimation procedure, all the
copulas in the subsequent trees are assumed to be product copulas. A
model selection procedure based on either AIC \citep[Akaike
information criterion;][]{Akaike1974AIC} or BIC \citep[Bayesian
information criterion;][]{Schwarz1978BIC} is applied to detect the
required number of trees. This procedure expands the tree~$T_{j+1}$ if
the value of the information criterion calculated up to the
tree~$T_{j+1}$ is smaller than the value obtained up to the previous
tree; otherwise, the vine is truncated at the tree~$T_{j}$. At this
point, it is important to note that the algorithm presented in
\cite{Salinas-Gutierrez2010DvineEDA} also uses a D-vine. In this
algorithm only normal copulas are fitted in the first two trees and
conditional independence is assumed in the rest of the trees, i.e.,
the D-vine is always truncated at the second tree.

The implementation of CVEDA and DVEDA included in \pkg{copulaedas}
uses by default the truncation procedure based on AIC and the
candidate copulas normal, $t$, Clayton, Frank and Gumbel. The
parameters of all copulas but the $t$~copula are estimated using the
method of moments. For the $t$~copula, the correlation coefficient is
computed as in the normal copula, and the degrees of freedom are
estimated by maximum likelihood with the correlation parameter fixed
\citep{Demarta2005tCopula}.

\section[Implementation in R]{Implementation in \proglang{R}\label{sec:implementation}}

According to the review presented by \citet{Santana2011EDAImplementations}, 
the approach followed for the implementation of EDA software currently available
through the Internet can be classified into three categories:
(1)~implementation of a single EDA, (2)~independent implementation of
multiple EDAs, and (3)~common modular implementation of multiple
EDAs. In our opinion, the third approach offers greater flexibility
for the EDA community.  In these modular implementations, the EDA
components (e.g., learning and sampling methods) are independently
programmed by taking advantage of the common schema shared by most
EDAs. This modularity allows the creation and validation of new EDA
proposals that combine different components, and promotes code
factorization. Additionally, as the EDAs are grouped under the same
framework, it facilitates performing empirical studies to compare the
behavior of different algorithms.  Existing members of this class are
\pkg{ParadisEO} \citep{Cahon2004ParadisEO, ParadisEO}, \pkg{LiO}
\citep{LiO, Mateo2007LiO}, \pkg{Mateda-2.0} 
\citep{Mateda,Santana2010Mateda20} and now \pkg{copulaedas}.

The implementation of \pkg{copulaedas} follows an object-oriented
design inspired by the \pkg{Mateda-2.0} toolbox for \proglang{MATLAB}
\citep{MATLAB}.  EDAs implemented in the package are represented by
\proglang{S}4~classes \citep{Chambers2008RProgramming} with generic
functions for their main steps. The base class of EDAs in the package
is `\code{EDA}', which has two slots: \code{name} and
\code{parameters}. The \code{name} slot stores a character string with
the name of the EDA and it is used by the \code{show} method to print
the name of the algorithm when it is called with an `\code{EDA}'
instance as argument. The \code{parameters} slot stores all the 
EDA parameters in a list.

In \pkg{copulaedas}, each step of the general procedure of an EDA
outlined in Algorithm~\ref{alg:eda-procedure} is represented by
a generic function that expects an `\code{EDA}' instance as its first
argument. Table~\ref{tab:generic-functions} shows a description
of these functions and their default methods. The help page of these
generic functions in the documentation of \pkg{copulaedas} contains
information about their arguments, return value, and methods
already implemented in the package.

\begin{table}[t!]
\centering
\begin{tabular}{lp{4.7in}l}
\hline 
Generic function & Description\\
\hline 
\code{edaSeed} & \emph{Seeding method. }The default method \code{edaSeedUniform} generates
the values of each variable in the initial population from a continuous
uniform distribution.\\
\code{edaOptimize} & \emph{Local optimization method}. The use of a local optimization
method is disabled by default.\\
\code{edaSelect} & \emph{Selection method}. The default method \code{edaSelectTruncation}
implements truncation selection.\\
\code{edaLearn} & \emph{Learning method}. No default method.\\
\code{edaSample} & \emph{Sampling method}. No default method.\\
\code{edaReplace} & \emph{Replacement method}. The default method \code{edaReplaceComplete}
\mbox{completely} replaces the current population with the new population. \\
\code{edaReport} & \emph{Reporting method}. Reporting progress information is disabled
by \mbox{default}.\\
\code{edaTerminate} & \emph{Termination method}. The default method \code{edaTerminateMaxGen}
ends the execution of the algorithm after a maximum number of generations.\\
\hline 
\end{tabular}
\caption{Description of the generic functions that implement the steps of the
general procedure of an EDA outlined in Algorithm~\ref{alg:eda-procedure}
and their default methods.\label{tab:generic-functions}}
\end{table}

The generic functions and their methods that implement the steps of an
EDA look at the \code{parameters} slot of the `\code{EDA}' instance
received as first argument for the values of the parameters that
affect their behavior. Only named members of the list must be used and
reasonable default values should be assumed when a certain component
is missing. The help page of each generic function describes the
members of the list in the \code{parameters} slot interpreted by each
function and their default values.

The \code{edaRun} function implements the Algorithm~\ref{alg:eda-procedure}
by linking together the generic functions for each step. This function
expects four arguments: the `\code{EDA}' instance, the objective function
and two vectors specifying the lower and upper bounds of the variables
of the objective function. The length of the vectors with the lower
and upper bounds should be the same, since it determines the number
of variables of the objective function. When \code{edaRun} is called,
it runs the main loop of the EDA until the call to the \code{edaTerminate}
generic function returns \code{TRUE}. Then, the function returns
an instance of the `\code{EDAResult}' class that encapsulates the results
of the algorithm. A description of the slots of this class is given
in Table~\ref{tab:EDAResult-slots}.

\begin{table}[t!]
\centering
\begin{tabular}{ll}
\hline 
Slot & Description\\
\hline 
\code{eda} & `\code{EDA}' instance.\\
\code{f} & Objective function.\\
\code{lower} & Lower bounds of the variables of the objective function.\\
\code{upper} & Upper bounds of the variables of the objective function.\\
\code{numGens} & Total number of generations.\\
\code{fEvals} & Total number of evaluations of the objective function.\\
\code{bestSol} & Best solution.\\
\code{bestEval} & Evaluation of the best solution.\\
\code{cpuTime} & Run time of the algorithm in seconds.\\
\hline 
\end{tabular}
\caption{Description of the slots of the `\code{EDAResult}' class.\label{tab:EDAResult-slots}}
\end{table}

Two subclasses of `\code{EDA}' are already defined in \pkg{copulaedas}:
`\code{CEDA}', that represents EDAs based on multivariate copulas;
and `\code{VEDA}', that represents vine-based EDAs. The implementation
of UMDA, GCEDA, CVEDA and DVEDA relies on the \pkg{copula}~\citep{copulaRPackage}, 
\pkg{vines}~\citep{vinesRPackage}, \pkg{mvtnorm}~\citep{mvtnormRPackage,Genz2009NormalAndTProbabilities},
and \pkg{truncnorm}~\citep{truncnormRPackage} \proglang{R} packages.
These packages implement the techniques for the estimation and simulation
of the probabilistic models used in these EDAs.

\section[Using copulaedas]{Using \pkg{copulaedas}\label{sec:examples}}

In this section, we illustrate how to use \pkg{copulaedas} through
several examples. To begin with, we show how to run the EDAs included
in the package. Next, we continue with the implementation of a new EDA
by using the functionalities provided by the package, and finally we
show how to perform an empirical study to compare the behavior of a
group of EDAs on benchmark functions and a real-world problem.

The two well-known test problems Sphere and Summation Cancellation
are used as the benchmark functions. The functions \code{fSphere}
and \code{fSummationCancellation} implement these problems in terms 
of a vector $\boldsymbol{x}=(x_{1},\ldots,x_{n})$ according to
\begin{align*}
f_{\textrm{Sphere}}(\boldsymbol{x})&=\sum_{i=1}^{n}x_{i}^{2},\\
f_{\textrm{Summation\ensuremath{\,}Cancellation}}(\boldsymbol{x})&=\frac{1}{10^{-5}+\sum_{i=1}^{n}|y_{i}|},\quad \, y_{1}=x_{1},\, y_{i}=y_{i-1}+x_{i}.
\end{align*}
Sphere is a minimization problem and Summation Cancellation is 
originally a maximization problem but it is defined in the package 
as a minimization problem. Sphere has its global optimum at 
$\boldsymbol{x}=(0,\ldots,0)$ with evaluation zero and Summation 
Cancellation at $\boldsymbol{x}=(0,\ldots,0)$ with evaluation 
$-10^{5}$. For a description of the characteristics of these 
functions see \cite{Bengoetxea2002ExperimentalResultsEDAs} and
\cite{Bosman2006RealValuedEDAs}.

The results presented in this section were obtained using \proglang{R}
version~3.1.0 with \pkg{copulaedas} version~1.4.0, \pkg{copula}
version~0.999-8, \pkg{vines} version~1.1.0, \pkg{mvtnorm} version~0.9-99991, 
and \pkg{truncnorm} version~1.0-7. Computations were performed on a 64-bit 
Linux machine with an Intel(R) Core(TM)2 Duo 2.00~GHz processor. In the rest 
of this section, we assume \pkg{copulaedas} has been loaded. This
can be attained by running the following command:
\begin{CodeChunk}
\begin{CodeInput}
R> library("copulaedas")
\end{CodeInput}
\end{CodeChunk}

\subsection{Running the EDAs included in the package}

We begin by illustrating how to run the EDAs based on copulas
implemented in \pkg{copulaedas}. As an example, we execute GCEDA to
optimize Sphere in five dimensions. Before creating a new instance of
the `\code{CEDA}' class for EDAs based on multivariate copulas, we set
up the generic functions for the steps of the EDA according to the
expected behavior of GCEDA. The termination criterion is either to
find the optimum of the objective function or to reach a maximum
number of generations.  That is why we set the method for the
\code{edaTerminate} generic function to a combination of the functions
\code{edaTerminateEval} and \code{edaTerminateMaxGen} through the
auxiliary function \code{edaTerminateCombined}.
\begin{CodeChunk}
\begin{CodeInput}
R> setMethod("edaTerminate", "EDA",
+    edaTerminateCombined(edaTerminateEval, edaTerminateMaxGen))
\end{CodeInput}
\end{CodeChunk}
The method for the \code{edaReport} generic function is set to \code{edaReportSimple}
to make the algorithm print progress information at each generation.
This function prints one line at each iteration of the EDA with the
minimum, mean and standard deviation of the evaluation of the solutions
in the current population.
\begin{CodeChunk}
\begin{CodeInput}
R> setMethod("edaReport", "EDA", edaReportSimple)
\end{CodeInput}
\end{CodeChunk}
Note that these methods were set for the base class `\code{EDA}' and
therefore they will be inherited by all subclasses. Generally, we find
it convenient to define methods of the generic functions that
implement the steps of the EDA for the base class, except when
different subclasses should use different methods.

The auxiliary function `\code{CEDA}' can be used to create instances of 
the class with the same name. All the arguments of the function are 
interpreted as parameters of the EDA to be added as members of the list 
in the \code{parameters} slot of the new instance. An instance of 
`\code{CEDA}' corresponding to GCEDA using empirical marginal 
distributions smoothed with normal kernels can be created as follows:
\begin{CodeChunk}
\begin{CodeInput}
R> gceda <- CEDA(copula = "normal", margin = "kernel", popSize = 200, 
+    fEval = 0, fEvalTol = 1e-6, maxGen = 50)
R> gceda@name <- "Gaussian Copula Estimation of Distribution Algorithm"
\end{CodeInput}
\end{CodeChunk}
The methods that implement the generic functions \code{edaLearn}
and \code{edaSample} for `\code{CEDA}' instances expect three parameters.
The \code{copula} parameter specifies the multivariate copula and
it should be set to \code{"normal"} for GCEDA. The marginal distributions
are determined by the value of \code{margin} and all EDAs implemented
in the package use this parameter for the same purpose. As \code{margin}
is set to \code{"kernel"}, the algorithm will look for three functions
named \code{fkernel}, \code{pkernel} and \code{qkernel} already
defined in the package to fit the parameters of the margins and to
evaluate the distribution and quantile functions, respectively. The
\code{fkernel} function computes the bandwidth parameter of the normal
kernel according to the rule-of-thumb of \citet{Silverman1986DensityEstimation}
and \code{pkernel} implements the empirical cumulative distribution
function. The quantile function is evaluated following the procedure
described in \cite{Azzalini1981DistributionKernel}. The \code{popSize}
parameter determines the population size while the rest of the arguments
of \code{CEDA} are parameters of the functions that implement the
termination criterion.

Now, we can run GCEDA by calling \code{edaRun}. The lower and upper
bounds of the variables are set so that the values of the variables in
the optimum of the function are located at 25\% of the interval.  It
was shown in \cite{Arderi2007BachelorThesis} and
\cite{Soto2014CopulaVineEDA} that the use of empirical marginal
distributions smoothed with normal kernels improves the behavior of
GCEDA when the initial population is generated asymmetrically with
respect to the optimum of the function.
\begin{CodeChunk}
\begin{CodeInput}
R> set.seed(12345)
R> result <- edaRun(gceda, fSphere, rep(-300, 5), rep(900, 5))
\end{CodeInput}
\begin{CodeOutput}
  Generation      Minimum         Mean    Std. Dev. 
           1 1.522570e+05 1.083606e+06 5.341601e+05 
           2 2.175992e+04 5.612769e+05 3.307403e+05 
           3 8.728486e+03 2.492247e+05 1.496334e+05 
           4 4.536507e+03 1.025119e+05 5.829982e+04 
           5 5.827775e+03 5.126260e+04 2.983622e+04 
           6 2.402107e+03 2.527349e+04 1.430142e+04 
           7 9.170485e+02 1.312806e+04 6.815822e+03 
           8 4.591915e+02 6.726731e+03 4.150888e+03 
           9 2.448265e+02 3.308515e+03 1.947486e+03 
          10 7.727107e+01 1.488859e+03 8.567864e+02 
          11 4.601731e+01 6.030030e+02 3.529036e+02 
          12 8.555769e+00 2.381415e+02 1.568382e+02 
          13 1.865639e+00 1.000919e+02 6.078611e+01 
          14 5.157326e+00 4.404530e+01 2.413589e+01 
          15 1.788793e+00 2.195864e+01 1.136284e+01 
          16 7.418832e-01 1.113184e+01 6.157461e+00 
          17 6.223596e-01 4.880880e+00 2.723950e+00 
          18 4.520045e-02 2.327805e+00 1.287697e+00 
          19 6.981399e-02 1.123582e+00 6.956201e-01 
          20 3.440069e-02 5.118243e-01 2.985175e-01 
          21 1.370064e-02 1.960786e-01 1.329600e-01 
          22 3.050774e-03 7.634156e-02 4.453917e-02 
          23 1.367716e-03 3.400907e-02 2.056747e-02 
          24 7.599946e-04 1.461478e-02 8.861180e-03 
          25 4.009605e-04 6.488932e-03 4.043431e-03 
          26 1.083879e-04 2.625759e-03 1.618058e-03 
          27 8.441887e-05 1.079075e-03 5.759307e-04 
          28 3.429462e-05 5.077934e-04 3.055568e-04 
          29 1.999004e-05 2.232605e-04 1.198675e-04 
          30 1.038719e-05 1.104123e-04 5.888948e-05 
          31 6.297005e-06 5.516721e-05 2.945027e-05 
          32 1.034002e-06 2.537823e-05 1.295004e-05 
          33 8.483830e-07 1.332463e-05 7.399488e-06 
\end{CodeOutput}
\end{CodeChunk}
The \code{result} variable contains an instance of the
`\code{EDAResult}' class. The \code{show} method prints the results of
the execution of the algorithm.
\begin{CodeChunk}
\begin{CodeInput}
R> show(result)
\end{CodeInput}
\begin{CodeOutput}
Results for Gaussian Copula Estimation of Distribution Algorithm
Best function evaluation    8.48383e-07 
No. of generations          33 
No. of function evaluations 6600 
CPU time                    7.895 seconds 
\end{CodeOutput}
\end{CodeChunk}
Due to the stochastic nature of EDAs, it is often useful to analyze
a sequence of independent runs to ensure reliable results. The \code{edaIndepRuns}
function supports performing this task. To avoid generating lot of
unnecessary output, we first disable reporting progress information
on each generation by setting \code{edaReport} to \code{edaReportDisabled}
and then we invoke the \code{edaIndepRuns} function to perform 30
independent runs of GCEDA.
\begin{CodeChunk}
\begin{CodeInput}
R> setMethod("edaReport", "EDA", edaReportDisabled)
R> set.seed(12345)
R> results <- edaIndepRuns(gceda, fSphere, rep(-300, 5), rep(900, 5), 30)
\end{CodeInput}
\end{CodeChunk}
The return value of the \code{edaIndepRuns} function is an instance of
the `\code{EDAResults}' class. This class is simply a wrapper for a
list with instances of `\code{EDAResult}' as members that contain the
results of an execution of the EDA. A \code{show} method for
`\code{EDAResults}' instances prints a table with all the results.
\begin{CodeChunk}
\begin{CodeInput}
R> show(results)
\end{CodeInput}
\begin{CodeOutput}
       Generations Evaluations Best Evaluation CPU Time
Run 1           33        6600    8.483830e-07    7.583
Run 2           38        7600    4.789448e-08    8.829
Run 3           36        7200    4.798364e-07    8.333
Run 4           37        7400    9.091651e-07    8.772
Run 5           34        6800    5.554465e-07    7.830
Run 6           35        7000    3.516341e-07    8.071
Run 7           35        7000    9.325531e-07    8.106
Run 8           35        7000    6.712550e-07    8.327
Run 9           36        7200    8.725061e-07    8.283
Run 10          37        7400    2.411458e-07    8.565
Run 11          36        7200    4.291725e-07    8.337
Run 12          35        7000    7.245520e-07    8.313
Run 13          37        7400    2.351322e-07    8.538
Run 14          36        7200    8.651248e-07    8.320
Run 15          34        6800    9.422646e-07    7.821
Run 16          35        7000    9.293726e-07    8.333
Run 17          36        7200    6.007390e-07    8.274
Run 18          38        7600    3.255231e-07    8.763
Run 19          36        7200    6.012969e-07    8.353
Run 20          35        7000    5.627017e-07    8.296
Run 21          36        7200    5.890752e-07    8.259
Run 22          35        7000    9.322505e-07    8.067
Run 23          35        7000    4.822349e-07    8.084
Run 24          34        6800    7.895408e-07    7.924
Run 25          36        7200    6.970180e-07    8.519
Run 26          34        6800    3.990247e-07    7.808
Run 27          35        7000    8.876874e-07    8.055
Run 28          33        6600    8.646387e-07    7.622
Run 29          36        7200    9.072113e-07    8.519
Run 30          35        7000    9.414666e-07    8.040
\end{CodeOutput}
\end{CodeChunk}
Also, the \code{summary} method can be used to generate a table with a
statistical summary of the results of the 30 runs of the algorithm.
\begin{CodeChunk}
\begin{CodeInput}
R> summary(results)
\end{CodeInput}
\begin{CodeOutput}
          Generations Evaluations Best Evaluation  CPU Time
Minimum     33.000000   6600.0000    4.789448e-08 7.5830000
Median      35.000000   7000.0000    6.841365e-07 8.2895000
Maximum     38.000000   7600.0000    9.422646e-07 8.8290000
Mean        35.433333   7086.6667    6.538616e-07 8.2314667
Std. Dev.    1.250747    250.1494    2.557100e-07 0.3181519
\end{CodeOutput}
\end{CodeChunk}

\subsection[Implementation of a new EDA based on
copulas]{Implementation of a new EDA based on
  copulas\label{sec:example-new-eda}}

In this section we illustrate how to use \pkg{copulaedas} to implement
a new EDA based on copulas. As an example, we consider the extension
of MIMIC for continuous domains proposed in
\cite{Salinas-Gutierrez2009CopulasEDAs}.  Similarly to MIMIC, this
extension learns a chain dependence structure, but it uses bivariate
copulas instead of bivariate normal distributions.  The chain
dependence structure is similar to a D-vine truncated at the first
tree, i.e., a D-vine where independence is assumed for all the trees
but the first. Two instances of the extension of MIMIC based on
copulas were presented in \cite{Salinas-Gutierrez2009CopulasEDAs},
one uses bivariate normal copulas while the other uses bivariate Frank
copulas. In this article, the algorithm will be denoted as
Copula~MIMIC.

Since the algorithm in question matches the general schema of an EDA
presented in Algorithm~\ref{alg:eda-procedure}, only the functions
corresponding to the learning and simulation steps have to be
implemented.  The first step in the implementation of a new EDA is to
define a new \proglang{S}4~class that inherits from `\code{EDA}' to
represent the algorithm. For convenience, we also define an auxiliary
function \code{CopulaMIMIC} that can be used to create new instances
of this class.
\begin{CodeChunk}
\begin{CodeInput}
R> setClass("CopulaMIMIC", contains = "EDA",
+    prototype = prototype(name = "Copula MIMIC"))
R> CopulaMIMIC <- function(...) 
+    new("CopulaMIMIC", parameters = list(...))
\end{CodeInput}
\end{CodeChunk}
Copula~MIMIC models the marginal distributions with the beta distribution.
A linear transformation is used to map the sample of the variables
in the selected population into the $(0,1)$ interval to match the
domain of definition of the beta distribution. Note that, since the
copula is scale-invariant, this transformation does not affect the
dependence between the variables. To be consistent with the margins
already implemented in \pkg{copulaedas}, we define three functions
with the common suffix \code{betamargin} and the prefixes \code{f},
\code{p} and \code{q} to fit the parameters of the margins and for
the evaluation of the distribution and quantile functions, respectively.
By following this convention, the algorithms already implemented in
the package can use beta marginal distributions by setting the \code{margin}
parameter to \code{"betamargin"}.
\begin{CodeChunk}
\begin{CodeInput}
R> fbetamargin <- function(x, lower, upper) {
+    x <- (x - lower) / (upper - lower)
+    loglik <- function(s) sum(dbeta(x, s[1], s[2], log = TRUE))
+    s <- optim(c(1, 1), loglik, control = list(fnscale = -1))$par
+    list(lower = lower, upper = upper, a = s[1], b = s[2])
+  }
R> pbetamargin <- function(q, lower, upper, a, b) {
+    q <- (q - lower) / (upper - lower)
+    pbeta(q, a, b)
+  }
R> qbetamargin <- function(p, lower, upper, a, b) {
+    q <- qbeta(p, a, b)
+    lower + q * (upper - lower)
+  }
\end{CodeInput}
\end{CodeChunk}
The `\code{CopulaMIMIC}' class inherits methods for the generic functions
that implement all the steps of the EDA except learning and sampling.
To complete the implementation of the algorithm, we must define the
estimation and simulation of the probabilistic model as methods for
the generic functions \code{edaLearn} and \code{edaSample}, respectively.

The method for \code{edaLearn} starts with the estimation of the
parameters of the margins and the transformation of the selected
population to uniform variables in $(0,1)$. Then, the mutual
information between all pairs of variables is calculated through the
copula entropy \citep{Davy2003CopulaEntropy}.  To accomplish this, the
parameters of each possible bivariate copula are estimated by the
method of maximum likelihood using the value obtained through the
method of moments as an initial approximation.  To determine the chain
dependence structure, a permutation of the variables that maximizes
the pairwise mutual information must be selected but, since this is a
computationally intensive task, a greedy algorithm is used to compute
an approximate solution
\citep{DeBonet1997MIMIC,Larranaga1999GaussianMIMIC}.  Finally, the
method for \code{edaLearn} returns a list with three components that
represents the estimated probabilistic model: the parameters of the
marginal distributions, the permutation of the variables, and the
copulas in the chain dependence structure.
\begin{CodeChunk}
\begin{CodeInput}
R> edaLearnCopulaMIMIC <- function(eda, gen, previousModel,
+      selectedPop, selectedEval, lower, upper) {
+    margin <- eda@parameters$margin
+    copula <- eda@parameters$copula
+    if (is.null(margin)) margin <- "betamargin"
+    if (is.null(copula)) copula <- "normal"
+    fmargin <- get(paste("f", margin, sep = ""))
+    pmargin <- get(paste("p", margin, sep = ""))
+    copula <- switch(copula, 
+      normal = normalCopula(0), frank = frankCopula(0))
+    n <- ncol(selectedPop)
+    # Estimate the parameters of the marginal distributions.
+    margins <- lapply(seq(length = n),
+      function(i) fmargin(selectedPop[, i], lower[i], upper[i]))
+    uniformPop <- sapply(seq(length = n), function(i) do.call(pmargin,
+      c(list(selectedPop[ , i]), margins[[i]])))
+    # Calculate pairwise mutual information by using copula entropy.
+    C <- matrix(list(NULL), nrow = n, ncol = n)
+    I <- matrix(0, nrow = n, ncol = n)
+    for (i in seq(from = 2, to = n)) {
+      for (j in seq(from = 1, to = i - 1)) {
+        # Estimate the parameters of the copula.
+        data <- cbind(uniformPop[, i], uniformPop[, j])
+        startCopula <- fitCopula(copula, data, method = "itau",
+          estimate.variance = FALSE)@copula
+        C[[i, j]] <- tryCatch(
+          fitCopula(startCopula, data, method = "ml",
+            start = startCopula@parameters, 
+            estimate.variance = FALSE)@copula,
+            error = function(error) startCopula)
+        # Calculate mutual information.
+        if (is(C[[i, j]], "normalCopula")) {
+          I[i, j] <- -0.5 * log(1 - C[[i, j]]@parameters^2)
+        } else {
+          u <- rcopula(C[[i, j]], 100)
+          I[i, j] <- sum(log(dcopula(C[[i, j]], u))) / 100
+        }
+        C[[j, i]] <- C[[i, j]]
+        I[j, i] <- I[i, j]
+      }
+    }
+    # Select a permutation of the variables.
+    perm <- as.vector(arrayInd(which.max(I), dim(I)))
+    copulas <- C[perm[1], perm[2]]
+    I[perm, ] <- -Inf
+    for (k in seq(length = n - 2)) {
+      ik <- which.max(I[, perm[1]])
+      perm <- c(ik, perm)
+      copulas <- c(C[perm[1], perm[2]], copulas)
+      I[ik, ] <- -Inf     
+    }
+    list(margins = margins, perm = perm, copulas = copulas)
+  }
R> setMethod("edaLearn", "CopulaMIMIC", edaLearnCopulaMIMIC)
\end{CodeInput}
\end{CodeChunk}
The \code{edaSample} method receives the representation of the
probabilistic model returned by \code{edaLearn} as the \code{model}
argument. The generation of a new solution with $n$ variables starts
with the simulation of an $n$-dimensional vector $U$ having uniform
marginal distributions in $(0,1)$ and the dependence described by the
copulas in the chain dependence structure. The first step is to
simulate an independent uniform variable $U_{\pi_{n}}$ in $(0,1)$,
where $\pi_{n}$ denotes the variable in the position $n$ of the
permutation $\pi$ selected by the \code{edaLearn} method.  The rest of
the uniform variables are simulated conditionally on the previously
simulated variable by using the conditional copula
$C(U_{\pi_{k}}|U_{\pi_{k+1}})$, with $k=n-1,n-2,\ldots,1$. Finally,
the new solution is determined through the evaluation of the beta
quantile functions and the application of the inverse of the linear
transformation.
\begin{CodeChunk}
\begin{CodeInput}
R> edaSampleCopulaMIMIC <- function(eda, gen, model, lower, upper) {
+    popSize <- eda@parameters$popSize
+    margin <- eda@parameters$margin
+    if (is.null(popSize)) popSize <- 100
+    if (is.null(margin)) margin <- "betamargin"
+    qmargin <- get(paste("q", margin, sep = ""))
+    n <- length(model$margins)
+    perm <- model$perm
+    copulas <- model$copulas
+    # Simulate the chain structure with the copulas.
+    uniformPop <- matrix(0, nrow = popSize, ncol = n)
+    uniformPop[, perm[n]] <- runif(popSize)
+    for (k in seq(from = n - 1, to = 1)) {
+      u <- runif(popSize)
+      v <- uniformPop[, perm[k + 1]]
+      uniformPop[, perm[k]] <- hinverse(copulas[[k]], u, v)
+    }
+    # Evaluate the inverse of the marginal distributions.
+    pop <- sapply(seq(length = n), function(i) do.call(qmargin,
+        c(list(uniformPop[, i]), model$margins[[i]])))
+    pop 
+  }
R> setMethod("edaSample", "CopulaMIMIC", edaSampleCopulaMIMIC)
\end{CodeInput}
\end{CodeChunk}
The code fragments given above constitute the complete implementation
of Copula~MIMIC. As it was illustrated with GCEDA in the previous
section, the algorithm can be executed by creating an instance of
the `\code{CopulaMIMIC}' class and calling the \code{edaRun} function.

\subsection{Performing an empirical study on benchmark problems}

We now show how to use \pkg{copulaedas} to perform an empirical study
of the behavior of a group of EDAs based on copulas on benchmark problems.
The algorithms to be compared are UMDA, GCEDA, CVEDA, DVEDA and Copula~MIMIC.
The first three algorithms are included in \pkg{copulaedas} and the
fourth algorithm was implemented in Section~\ref{sec:example-new-eda}.
The two functions Sphere and Summation Cancellation described at the 
beginning of Section~\ref{sec:examples} are considered as benchmark 
problems in 10 dimensions. 

The aim of this empirical study is to assess the behavior of these
algorithms when only linear and independence relationships are considered.
Thus, only normal and product copulas are used in these EDAs. UMDA
and GCEDA use multivariate product and normal copulas, respectively.
CVEDA and DVEDA are configured to combine bivariate product and normal
copulas in the vines. Copula~MIMIC learns a chain dependence structure
with normal copulas. All algorithms use normal marginal distributions.
Note that in this case, GCEDA corresponds to EMNA and Copula~MIMIC
is similar to MIMIC for continuous domains. In the following code
fragment, we create class instances corresponding to these algorithms.
\begin{CodeChunk}
\begin{CodeInput}
R> umda <- CEDA(copula = "indep", margin = "norm")
R> umda@name <- "UMDA"
R> gceda <- CEDA(copula = "normal", margin = "norm")
R> gceda@name <- "GCEDA"
R> cveda <- VEDA(vine = "CVine", indepTestSigLevel = 0.01,    
+    copulas = c("normal"), margin = "norm")
R> cveda@name <- "CVEDA"
R> dveda <- VEDA(vine = "DVine", indepTestSigLevel = 0.01,
+    copulas = c("normal"), margin = "norm")
R> dveda@name <- "DVEDA"
R> copulamimic <- CopulaMIMIC(copula = "normal", margin = "norm") 
R> copulamimic@name <- "CopulaMIMIC"
\end{CodeInput}
\end{CodeChunk}
The initial population is generated using the default \code{edaSeed}
method, therefore, it is sampled uniformly in the real interval of
each variable. The lower and upper bounds of the variables are set so
that the values of the variables in the optimum of the function are
located in the middle of the interval.  We use the intervals
$[-600,600]$ in Sphere and $[-0.16,0.16]$ in Summation Cancellation. 
All algorithms use the default truncation selection method with a 
truncation factor of 0.3. Three termination criteria are combined 
using the \code{edaTerminateCombined} function: to find the 
global optimum of the function with a precision greater
than $10^{-6}$, to reach $300000$ function evaluations, or to loose
diversity in the population (i.e., the standard deviation of the
evaluation of the solutions in the population is less than
$10^{-8}$). These criteria are implemented in the functions
\code{edaTerminateEval}, \code{edaTerminateMaxEvals} and
\code{edaTerminateEvalStdDev}, respectively.

The population size of EDAs along with the truncation method determine
the sample available for the estimation of the search distribution.
An arbitrary selection of the population size could lead to misleading
conclusions of the results of the experiments. When the population
size is too small, the search distributions might not be accurately
estimated. On the other hand, the use of an excessively large population
size usually does not result in a better behavior of the algorithms
but certainly in a greater number of function evaluations. Therefore,
we advocate for the use of the critical population size when comparing
the performance of EDAs. The critical population size is the minimum
population size required by the algorithm to find the global optimum
of the function with a high success rate, e.g., to find the optimum
in 30 of 30 sequential independent runs.

An approximate value of the critical population size can be determined
empirically using a bisection method (see e.g., \citealp{Pelikan2005hBOA}
for a pseudocode of the algorithm). The bisection method begins with
an initial interval where the critical population size should be located
and discards one half of the interval at each step. This procedure
is implemented in the \code{edaCriticalPopSize} function. In the
experimental study carried out in this section, the initial interval
is set to $[50,2000]$. If the critical population size is not found
in this interval, the results of the algorithm with the population
size given by the upper bound are presented.

The complete empirical study consists of performing 30 independent
runs of every algorithm on every function using the critical population
size. We proceed with the definition of a list containing all algorithm-function
pairs.
\begin{CodeChunk}
\begin{CodeInput}
R> edas <- list(umda, gceda, cveda, dveda, copulamimic) 
R> fNames <- c("Sphere", "SummationCancellation")
R> experiments <- list()
R> for (eda in edas) {     
+    for (fName in fNames) {         
+      experiment <- list(eda = eda, fName = fName)         
+      experiments <- c(experiments, list(experiment))     
+    } 
+  }
\end{CodeInput}
\end{CodeChunk}
Now we define a function to process the elements of the \code{experiments}
list. This function implements all the experimental setup described
before. The output of \code{edaCriticalPopSize} and \code{edaIndepRuns}
is redirected to a different plain text file for each algorithm-function
pair.
\begin{CodeChunk}
\begin{CodeInput}
R> runExperiment <- function(experiment) {
+    eda <- experiment$eda
+    fName <- experiment$fName
+    # Objective function parameters.
+    fInfo <- list(
+      Sphere = list(lower = -600, upper = 600, fEval = 0),
+      SummationCancellation = list(lower = -0.16, upper = 0.16, 
+        fEval = -1e5))
+    lower <- rep(fInfo[[fName]]$lower, 10)
+    upper <- rep(fInfo[[fName]]$upper, 10)
+    f <- get(paste("f", fName, sep = ""))
+    # Configure termination criteria and disable reporting.
+    eda@parameters$fEval <- fInfo[[fName]]$fEval
+    eda@parameters$fEvalTol <- 1e-6
+    eda@parameters$fEvalStdDev <- 1e-8
+    eda@parameters$maxEvals <- 300000
+    setMethod("edaTerminate", "EDA",
+      edaTerminateCombined(edaTerminateEval, edaTerminateMaxEvals,
+        edaTerminateEvalStdDev))
+    setMethod("edaReport", "EDA", edaReportDisabled)
+    sink(paste(eda@name, "_", fName, ".txt", sep = ""))
+    # Determine the critical population size.
+    set.seed(12345)
+    results <- edaCriticalPopSize(eda, f, lower, upper,
+      eda@parameters$fEval, eda@parameters$fEvalTol, lowerPop = 50,
+      upperPop = 2000, totalRuns = 30, successRuns = 30,
+      stopPercent = 10, verbose = TRUE)
+    if (is.null(results)) {
+      # Run the experiment with the largest population size, if the
+      # critical population size was not found.
+      eda@parameters$popSize <- 2000
+      set.seed(12345)
+      edaIndepRuns(eda, f, lower, upper, runs = 30, verbose = TRUE)
+    }
+    sink(NULL)
+  }
\end{CodeInput}
\end{CodeChunk}
We can run all the experiments by calling \code{runExperiment} for
each element of the list.
\begin{CodeChunk}
\begin{CodeInput}
R> for (experiment in experiments) runExperiment(experiment)
\end{CodeInput}
\end{CodeChunk}
Running the complete empirical study sequentially is a computationally 
demanding operation. If various processing units are available, it can 
be speeded up significantly by running the experiments in parallel. 
The \pkg{snow} package \citep{snowRPackage} offers a great platform 
to achieve this purpose, since it provides a high-level interface 
for using a cluster of workstations for parallel computations
in \proglang{R}. The functions \code{clusterApply} or \code{clusterApplyLB}
can be used to call \code{runExperiment} for each element of the
\code{experiments} list in parallel, with minimal modifications to
the code presented here.

A summary of the results of the algorithms with the critical population size 
is shown in Table~\ref{tab:benchmark-results}. Overall, the five algorithms are 
able to find the global optimum of Sphere in all the 30 independent 
runs with similar function values but only GCEDA, CVEDA and DVEDA 
optimize Summation Cancellation. In the rest of this section 
we provide some comments about results of the algorithms on each 
function.

\begin{table}[t!]
\centering
\begin{tabular}{@{}lccc@{\hspace{15pt}}c@{\hspace{18pt}}c@{}}
\hline 
Algorithm & Success & Pop. & Evaluations & Best Evaluation & CPU Time\\
\hline 
\multicolumn{6}{@{}l}{\emph{Sphere:}}\\
UMDA & 30/30 & 81 & \hspace{-0.5em}3788.1~$\pm$~99.0 & $6.7$e$-07$~$\pm$~$2.0$e$-07$ & 0.2~$\pm$~0.0\\
GCEDA & 30/30 & 310 & \hspace{-0.5em}13102.6~$\pm$~180.8 & $6.8$e$-07$~$\pm$~2.0e$-07$ & 0.4~$\pm$~0.0\\
CVEDA & 30/30 & 104 & \hspace{-0.5em}4804.8~$\pm$~99.9 & 5.9e$-07$~$\pm$~1.6e$-07$ & 4.5~$\pm$~0.8\\
DVEDA & 30/30 & 111 & 5080.1~$\pm$~111.6 & 6.5e$-07$~$\pm$~2.3e$-07$ & 4.4~$\pm$~0.8\\
Copula~MIMIC & 30/30 & 172 & 7441.8~$\pm$~127.2 & 6.8e$-07$~$\pm$~1.9e$-07$ & \hspace{-0.5em}65.6~$\pm$~1.2\\
\hline 
\multicolumn{6}{@{}l}{\emph{Summation Cancellation:}}\\
UMDA & \hspace{0.5em}0/30 & 2000 & \hspace{-2em}300000.0~$\pm$~0.0 & \hspace{-0.75em}$-7.2$e+02~$\pm$~4.5e+02 & \hspace{-0.5em}32.5~$\pm$~0.3\\
GCEDA & 30/30 & 325 & \hspace{-0.5em}38913.3~$\pm$~268.9 & \hspace{-0.75em}$-1.0$e+05~$\pm$~1.0e$-07$ & 4.2~$\pm$~0.1\\
CVEDA & 30/30 & 294 & 41228.6~$\pm$~1082.7 & \hspace{-0.75em}$-1.0$e+05~$\pm$~1.0e$-07$ & \hspace{-1em}279.1~$\pm$~5.1\\
DVEDA & 30/30 & 965 & \hspace{-0.5em}117022.3~$\pm$~1186.8 & \hspace{-0.75em}$-1.0$e+05~$\pm$~1.1e$-07$ & \hspace{-1em}1617.5~$\pm$~39.1\\
Copula~MIMIC & \hspace{0.5em}0/30 & 2000 & \hspace{-2em}300000.0~$\pm$~0.0 & \hspace{-0.75em}$-4.2$e+04~$\pm$~3.5e+04 & \hspace{-1em}1510.6~$\pm$~54.9\\
\hline 
\end{tabular}
\caption{Summary of the results obtained in 30 independent runs of UMDA, GCEDA,
CVEDA, DVEDA and Copula~MIMIC in the 10-dimensional Sphere (top) and
Summation Cancellation (bottom). Pop.~denotes Population.\label{tab:benchmark-results}}
\end{table}

UMDA exhibits the best behavior in terms of the number of function
evaluations in Sphere. There are no strong dependences between the
variables of this function and the results suggest that considering
the marginal information is enough to find the global optimum efficiently.
The rest of the algorithms being tested require the calculation of
a greater number of parameters to represent the relationships between
the variables and hence larger populations are needed to compute them
reliably (\citealp{Soto2014CopulaVineEDA} illustrate this issue in
more detail with GCEDA). CVEDA and DVEDA do not assume a normal dependence
structure between the variables and for this reason are less affected
by this issue. The estimation procedure used by the vine-based algorithms
selects the product copula if there is not enough evidence of dependence.

Both UMDA and Copula~MIMIC fail to optimize Summation Cancellation.
A correct representation of the strong linear interactions between the
variables of this function seems to be essential to find the global
optimum. UMDA completely ignores this information by assuming
independence between the variables and it exhibits the worst
behavior. Copula~MIMIC reaches better fitness values than UMDA but
neither can find the optimum of the function. The probabilistic model
estimated by Copula~MIMIC cannot represent important dependences
necessary for the success of the optimization. The algorithms GCEDA, 
CVEDA and DVEDA do find the global optimum of Summation Cancellation. 
The results of GCEDA are slightly better than the ones of CVEDA and 
these two algorithms achieve much better results than DVEDA in 
terms of the number of function evaluations. The correlation matrix 
estimated by GCEDA can properly represent the multivariate linear 
interactions between the variables. The C-vine structure used in CVEDA,
on the other hand, provides a very good fit for the dependence structure 
between the variables of Summation Cancellation, given that it is possible 
to find a variable that governs the interactions in the sample 
(see \citealp{Gonzalez-Fernandez2011BachelorThesis} for more details).

The running time of Copula~MIMIC is considerably greater than the
running time of the other algorithms for all the functions. This
result is due to the use of a numerical optimization algorithm for the
estimation of the parameters of the copulas by maximum likelihood. In
the context of EDAs, where copulas are fitted at every generation, the
computational effort required to estimate the parameters of the
copulas becomes an important issue. As was illustrated with CVEDA and
DVEDA, using a method of moments estimation is a viable alternative to
maximum likelihood that requires much less CPU time. The empirical
investigation confirms the robustness of CVEDA and DVEDA in problems
with both weak and strong interactions between the variables. 
Nonetheless, the flexibility afforded by these algorithms comes with an 
increased running time when compared to UMDA or GCEDA, since the 
interactions between the variables have to be discovered during 
the learning step.

A general result of this empirical study is that copula-based EDAs
should use copulas other than the product only when there is evidence
of dependence. Otherwise, the EDA will require larger populations
and hence a greater number of function evaluations to accurately determine
the parameters of the copulas that correspond to independence.

\subsection{Solving the molecular docking problem}

Finally, we illustrate the use of \pkg{copulaedas} for solving a
so-called real-world problem. In particular, we use CVEDA and DVEDA to
solve an instance of the molecular docking problem, which is an
important component of protein-structure-based drug design.  From the
point of view of the computational procedure, it entails predicting
the geometry of a small ligand that binds to the active site of a
large macromolecular protein receptor. Protein-ligand docking remains
being a highly active area of research, since the algorithms for
exploring the conformational space and the scoring functions that have
been implemented so far have significant limitations
\citep{Warren2006CriticalAssessmentDockingPrograms}.

In our docking simulations, the protein is treated as a rigid body
while the ligand is fully flexible. Thus, a candidate solution represents
only the geometry of the ligand and it is encoded as a vector of real
values that represent its position, orientation and flexible torsion
angles. The first three variables of this vector represent the ligand
position in the three-dimensional space constrained to a box enclosing
the receptor binding site. The construction of this box is based on
the minimum and maximum values of the ligand coordinates in its crystal
conformation plus a padding distance of 5{\AA{}} added to
each main direction of the space. The remainder vector variables are
three Euler angles that represent the ligand orientation as a rigid
body and take values in the intervals $[0,2\pi]$, $[-\pi/2,\pi/2]$
and $[-\pi,\pi]$, respectively; and one additional variable restricted
to $[-\pi,\pi]$ for each flexible torsion angle of the ligand.

The semiempirical free-energy scoring function implemented as part
of the suite of automated docking tools \pkg{AutoDock}~4.2 \citep{Morris2009AutoDock4}
is used to evaluate each candidate ligand conformation. The overall
binding energy of a given ligand molecule is expressed as the sum
of the pairwise interactions between the receptor and ligand atoms
(intermolecular interaction energy), and the pairwise interactions
between the ligand atoms (ligand intramolecular energy). The terms
of the function consider dispersion/repulsion, hydrogen bonding, electrostatics,
and desolvation effects, all scaled empirically by constants determined
through a linear regression analysis. The aim of an optimization algorithm
performing the protein-ligand docking is to minimize the overall energy
value. Further details of the energy terms and how the function is
derived can be found in \cite{Huey2007AutoDockFunction}. 

Specifically, we consider as an example here the docking of the 2z5u
test system, solved by \mbox{X-ray} crystallography and available as part of
the Protein Data Bank \citep{Berman2000PDB}. The protein receptor is
lysine-specific histone demethylase~1 and the ligand is a \mbox{73-atom}
molecule (non-polar hydrogens are not counted) with 20 ligand torsions
in a box of 28{\AA{}}$\times$32{\AA{}}$\times$24{\AA{}}.  In order to
make it easier for the readers of the paper to reproduce this example,
the implementation of the \pkg{AutoDock}~4.2 scoring function in
\proglang{C} was extracted from the original program and it is
included (with unnecessary dependences removed) in the supplementary
material as \code{docking.c}. During the evaluation of the scoring
function, precalculated grid maps (one for each atom type present in
the ligand being docked) are used to make the docking calculations
fast. The result of these precalculations and related metadata for the
2z5u test system are contained in the attached ASCII file
\code{2z5u.dat}.

We make use of the system command \code{R CMD SHLIB} to build a shared
object for dynamic loading from the file \code{docking.c}. Next, we
integrate the created shared object into \proglang{R} using the
function \code{dyn.load} and load the precalculated grids using the
utility \proglang{C} function \code{docking_load} as follows.
\begin{CodeChunk}
\begin{CodeInput}
R> system("R CMD SHLIB docking.c")
R> dyn.load(paste("docking", .Platform$dynlib.ext, sep = ""))
R> .C("docking_load", as.character("2z5u.dat"))
\end{CodeInput}
\end{CodeChunk}
The docking of the 2z5u test system results in a minimization problem
with a total of 26 variables. Two vectors with the lower and upper
bounds of these variables are defined using utility functions provided
in \code{docking.c} to compute the bounds of the variables that
determine the position of the ligand and the total number of
torsions. For convenience, we also define an \proglang{R} wrapper
function \code{fDocking} for the \proglang{C} scoring function
\code{docking_score} provided in the compiled code that was loaded
into \proglang{R}.
\begin{CodeChunk}
\begin{CodeInput}
R> lower <- c(.C("docking_xlo", out = as.double(0))$out,
+    .C("docking_ylo", out = as.double(0))$out,
+    .C("docking_zlo", out = as.double(0))$out, 0, -pi/2, -pi,
+    rep(-pi, .C("docking_ntor", out = as.integer(0))$out))
R> upper <- c(.C("docking_xhi", out = as.double(0))$out,
+    .C("docking_yhi", out = as.double(0))$out,
+    .C("docking_zhi", out = as.double(0))$out, 2 * pi, pi/2, pi, 
+    rep(pi, .C("docking_ntor", out = as.integer(0))$out))
R> fDocking <- function(sol) 
+    .C("docking_score", sol = as.double(sol), out = as.double(0))$out
\end{CodeInput}
\end{CodeChunk}
CVEDA and DVEDA are used to solve the minimization problem, since they
are the most robust algorithms among the EDAs based on copulas
implemented in \pkg{copulaedas}. The parameters of these algorithms
are set to the values reported by \citet{Soto2012VEDAMD} in 
the solution of the 2z5u test system.  The
population size of CVEDA and DVEDA is set to 1400 and 1200,
respectively.  Both algorithms use the implementation of the truncated
normal marginal distributions
\citep{Johnson1994ContinuousUnivariateDistributions} provided by
\citet{truncnormRPackage} to satisfy the box constraints of the
variables. The termination criterion of both CVEDA and DVEDA is to
reach a maximum of 100 generations, since an optimum value of the
scoring function is not known. Instances of the `\code{VEDA}' class
that follow the description given above are created with the following
code.
\begin{CodeChunk}
\begin{CodeInput}
R> setMethod("edaTerminate", "EDA", edaTerminateMaxGen)
R> cveda <- VEDA(vine = "CVine", indepTestSigLevel = 0.01,
+    copulas = "normal", margin = "truncnorm", popSize = 1400, maxGen = 100)
R> dveda <- VEDA(vine = "DVine", indepTestSigLevel = 0.01,
+    copulas = "normal", margin = "truncnorm", popSize = 1200, maxGen = 100)
\end{CodeInput}
\end{CodeChunk}
Now we proceed to perform 30 independent runs of each algorithm using
the \code{edaIndepRuns} function. The arguments of this function
are the instances of the `\code{VEDA}' class corresponding to CVEDA
and DVEDA, the scoring function \code{fDocking}, and the vectors
\code{lower} and \code{upper} that determine the bounds of the variables.
\begin{CodeChunk}
\begin{CodeInput}
R> set.seed(12345)
R> cvedaResults <- edaIndepRuns(cveda, fDocking, lower, upper, runs = 30)
R> summary(cvedaResults)
\end{CodeInput}
\begin{CodeOutput}
          Generations Evaluations Best Evaluation  CPU Time
Minimum           100      140000      -30.560360 1421.9570
Median            100      140000      -29.387815 2409.0680
Maximum           100      140000      -23.076769 3301.5960
Mean              100      140000      -29.139028 2400.4166
Std. Dev.           0           0        1.486381  462.9442
\end{CodeOutput}
\begin{CodeInput}
R> set.seed(12345)
R> dvedaResults <- edaIndepRuns(dveda, fDocking, lower, upper, runs = 30)
R> summary(dvedaResults)
\end{CodeInput}
\begin{CodeOutput}
          Generations Evaluations Best Evaluation  CPU Time
Minimum           100      120000       -30.93501 1928.2030
Median            100      120000       -30.70019 3075.7960
Maximum           100      120000       -24.01502 4872.1820
Mean              100      120000       -30.01427 3053.9579
Std. Dev.           0           0         1.68583  695.0071
\end{CodeOutput}
\end{CodeChunk}
\begin{table}[!]
\centering
\begin{tabular}{@{}lccccc@{}}
\hline 
{Algorithm} & {Pop.} & {Evaluations} & {Lowest Energy} & {RMSD} & {CPU Time}\\
\hline 
CVEDA & 1400 & 140000.0~$\pm$~0.0 & $-$29.13~$\pm$~1.48 & 0.58~$\pm$~0.13 & 2400.4~$\pm$~462.9\\
DVEDA & 1200 & 120000.0~$\pm$~0.0 & $-$30.01~$\pm$~1.68 & 0.56~$\pm$~0.14 & 3053.9~$\pm$~695.0\\
\hline 
\end{tabular}
\par
\caption{Summary of the results obtained in 30 independent runs of CVEDA and
DVEDA for the docking of the 2z5u test system. Pop.~denotes 
Population.\label{tab:docking-results}}
\end{table}

\begin{figure}[t!]
\centering
\input{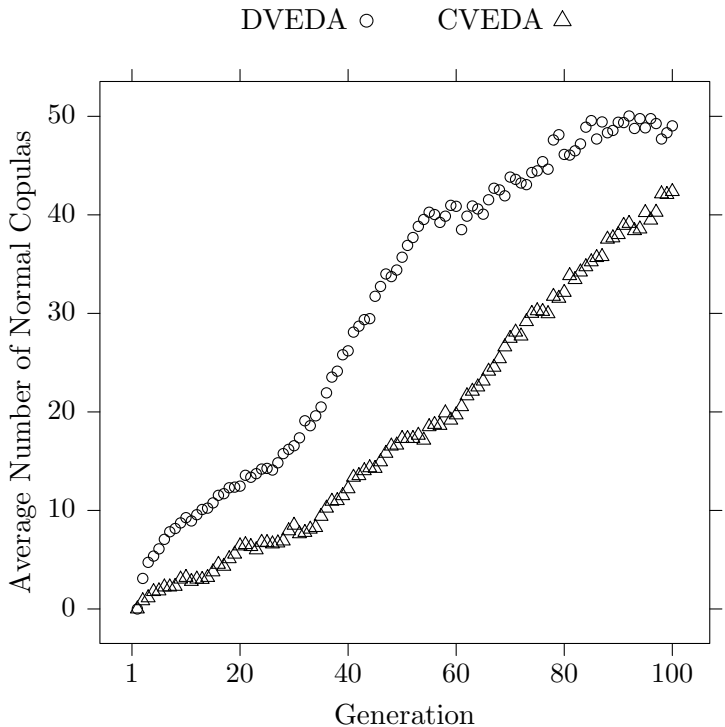}
\par
\caption{Average number of normal copulas selected at each generation
  of CVEDA and DVEDA in 30 independent runs for the docking of the
  2z5u test system. Since this is an optimization problem with 26
  variables, the C-vines and D-vines have a total of 325
  copulas.\label{fig:normal-copulas}}
\end{figure}

The results obtained in the docking of the 2z5u test system are
summarized in Table~\ref{tab:docking-results}. In addition to the
information provided by \code{edaIndepRuns}, we present a column for
the RMSD (root mean square deviation) between the coordinates of the
atoms in the experimental crystal structure and the predicted ligand
coordinates of the best solution found at each run. These values can
be computed for a particular solution using the \code{docking_rmsd}
function included in \code{docking.c}. The RMSD values serve as a
measure of the quality of the predicted ligand conformations when an
experimentally determined solution is known.

Generally, a structure with RMSD below 2{\AA{}} can be considered as
successfully docked. Therefore, both CVEDA and DVEDA achieve good
results when solving the 2z5u test system. DVEDA exhibits slightly
lower energy and RMSD values, and it requires a smaller population
size than CVEDA. On the other hand, DVEDA uses more CPU time than
CVEDA, a fact that might be related with more dependences being
encoded in the D-vines estimated by DVEDA than in the C-vines used by
CVEDA.  This situation is illustrated in
Figure~\ref{fig:normal-copulas}, which presents the average number of
normal copulas selected by CVEDA and DVEDA at each generation in the
30 independent runs. In both algorithms, the number of normal copulas
increases during the evolution, but DVEDA consistently selects more
normal copulas than CVEDA. Although the estimation procedure of the
C-vines in CVEDA intends to represent the strongest dependences in the
first tree, the constraint that only one variable can be connected to
all the others prevent strong correlations to be explicitly
encoded in the C-vines.

It is also worth noting that because of the use of the truncation
procedure based on AIC in both CVEDA and DVEDA, the number of
statistical tests required to estimate the vines was dramatically
reduced. The median number of vine trees fitted in the 30 independent
runs was 4 in CVEDA and 5 in DVEDA out of a total of 25 trees. The
interested reader is referred to \cite{Soto2012VEDAMD} for a deeper
study of the use of EDAs based on copulas for solving the molecular
docking problem.

\section[Concluding remarks]{Concluding remarks\label{sec:conclusions}}

We have developed \pkg{copulaedas} aiming at providing in a single
package an open-source implementation of EDAs based on copulas and
utility functions to study these algorithms. In this paper, we
illustrate how to run the copula-based EDAs implemented in the
package, how to implement new algorithms, and how to perform an
empirical study to compare copula-based EDAs on benchmark functions
and practical problems. We hope that these functionalities help the
research community to improve EDAs based on copulas by getting a
better insight of their strengths and weaknesses, and also help
practitioners to find new applications of these algorithms to
real-world problems.

\end{document}